\documentclass[conference]{IEEEtran}
\usepackage{cite}
\usepackage{amsmath,amssymb,amsfonts,graphicx}
\usepackage{algorithmic}
\usepackage{graphicx}
\usepackage{textcomp}
\usepackage{xcolor}
\usepackage{subfig}
\graphicspath{{imgs/}}
\usepackage{geometry}
\usepackage{multicol}
\def\BibTeX{{\rm B\kern-.05em{\sc i\kern-.025em b}\kern-.08em
    T\kern-.1667em\lower.7ex\hbox{E}\kern-.125emX}}
\begin{document}

\title{Autoencoder Watchdog Outlier Detection for Classifiers }

\author{\IEEEauthorblockN{Justin Bui}
\IEEEauthorblockA{\textit{Department of Electrical and Computer Engineering} \\
	\textit{Baylor University} \\
Waco, TX \\
Justin\_Bui@baylor.edu}
\and
\IEEEauthorblockN{Robert J Marks II, PhD}
\IEEEauthorblockA{\textit{Department of Electrical and Computer Engineering} \\
\textit{Baylor University}\\
Waco, TX \\
Robert\_Marks@baylor.edu}
}

\maketitle
\begin{abstract}
Neural networks have often been described as black boxes. A generic neural network trained to differentiate between kittens and puppies will classify a picture of a kumquat as a kitten or a puppy.  An autoencoder watchdog screens trained classifier/regression machine input candidates before processing, e.g.  to first test whether the neural network input is a puppy or a kitten. Preliminary results are presented using convolutional neural networks and convolutional autoencoder watchdogs using MNIST images.     

\end{abstract}

\section{Introduction}

Akin to principle component analysis \cite{Oja}, autoencoders can implicitly learn by the estimation of a lower dimensional manifold on which training data lives \cite{Thompson,Thompson2}. The feature space dimension is determined by the cardinality of the autoencoder's input and output. The dimension of the manifold is dictated by the size of the bottleneck layer (or waist) of the autoencoder. Representative test data presented to a properly trained autoencoder will generate an output similar to the input. 

More generally, the root mean square error (RSME) between the autoencoder input and output can be viewed as the rough distance measurement between the autoencoder input and the training data manifold in the feature space. For this reason, autoencoders can be used in novelty detection \cite{Guttormsson,Streifel,Thompson}. 

One could train a neural network on three outputs: kittens, puppies and all other images that are not kittens or puppies. One challenge to this approach is that the set of images that do not contain kittens or puppies is prohibitively large. Work done by Abbasi and DeVries \cite{Abbasi, DeVries} suggest strategies to work around the need for these large datasets. 

Use of an autoencoder watchdog is a more reasonable solution \cite{Streifel}, acting as a novelty (or anomaly) detector which protects the classifier neural network from fraudulent inputs. The autoencoder generates the manifold of data points that represent kittens and puppies. Any image lying far from the manifold is not a kitten or a puppy. 
	
A data point that lies close to the manifold need not be a kitten or a puppy. Another image may coincidently lie on the manifold. In anomaly detection, a flag raised by the autoencoder is therefore sufficient for detecting anomalies but is not necessary for detecting outliers.  

\section{Background}

Interest in, and the application of neural networks \cite{Reed} continues to expand at a rapid rate and cover a variety of tasks of varying complexities. Yadav et al present an excellent introduction to the history of neural networks \cite{Yadav}. Autoencoder neural networks are of particular interest in watchdog novelty detection. They have been used in a variety of different applications and may be implemented in a variety of different ways. For example, Baur \cite{Baur} has demonstrated anomaly detection in medical scans, whereas Alvernaz \cite{Alvernaz} explored the ability to visually analyze and learn to play complex videogames. Vu \cite{Vu} has investigated anomaly detection using adversarial autoencoders, while Lore \cite{Lore} has reported their use in low-light image enhancement applications. 

Autoencoders are useful for denoising various types of data, from udio  to medical images. Work done by Gondara \cite{Gondara} and Vincent \cite{Vincent} provide excellent examples of the these denoising techniques.  Autoencoders have been also been used in generative networks, as described by Mesheder \cite{Mescheder}. Most commonly used in generative adversarial networks, or GANs, autoencoders have shown remarkable capabilities in generating images from noise. Work done by Zhifei Zhang \cite{Zhang2}, Zijun Zhang \cite{Zhang3}, Huang \cite{Huang}, and Ranjan \cite{Ranjan} have shown some impressive generative capabilities across multiple spectrums, from grayscale digits to 3D face images. Work done by Luo \cite{Luo} demonstrates different techniques based on the combination of variational autoencoders (VAEs) and GANs. Work done by Lu \cite{Lu}, Xia \cite{Xia}, and Qi \cite{Qi} have demonstrated various applications beyond denoising and generation, highlighting the flexibility and useability of autoencoders.


As the artificial neural network field continues to grow and new tools continue to be developed, it is becoming easier to develop neural networks without a deep understanding of the driving principles. These new tools (eg TensorFlow, PyTorch, Keras, FastAI) lead to many neural networks being generally treated as {\em black boxes}. Our interest in these black boxes, as described by Alain et al \cite{Alain}, is less aimed at diving into the inner workings and attempting to demystify them, but rather to develop a technique that may be used with both existing and newly developed neural networks to address the uncertainty born of opaque neural networks. While there have been several attempts at diving in to the understanding of neural networks, such as the work done by Schartz \cite{Shwartz}, Zeiler \cite{Zeiler}, Martin \cite{Martin} and Markopoulos \cite{Markopoulos}, much of today's end products are assumed to be ``black boxes''.

 While there is no shortage of neural network structures and applications, our research focuses on {\em convolutional neural networks} (CNN's) and autoencoders. CNN's have demonstrated impressive performance in the classification and generation of data. As an example, Zhang \cite{Zhang} provides an excellent introduction to the concept of text understanding, paralleling the interpretation of hand written digits. Bhatnagar \cite{Bhatnagar} demonstrates the classification capabilities of CNNs on clothing items. Other work \cite{Ciresan,Tabik}  further details these capabilities while introducing new approaches for network design and performance optimization.

\section{The Neural Network Watchdog}
The {\em neural network watchdog} is a tool to determine a neural network's output validity. This is achieved by using the generative component of the autoencoder to reconstruct the input data and calculating a difference score. The difference score is then compared to a threshold that determines data validity. In this paper, we build on the use of autoencoders to create the generative component of the watchdog. For our differencing component,  the root mean square error (RMSE: the square root of the sum of the squares) is calculated and comparing against a fixed threshold. Below are descriptions of the  classification and watchdog autoencoder networks, the training and evaluation datasets, and the classifier and watchdog performance analysis.

\subsection{Network Structures}
For the viability study, both our classifier and autoencoder are CNN's. Work by LeCun \cite{LeCun}, Ng \cite{Ng}, and Meng \cite{Meng} provide an excellent foundation for designing such networks. The MNIST handwritten digit image dataset is used to train the neural network and its watchdog autoencoder.  

\subsubsection{The Autoencoder Watchdog}
\begin{figure}[h!]
	\begin{center}
		\includegraphics[width=.3\textwidth]{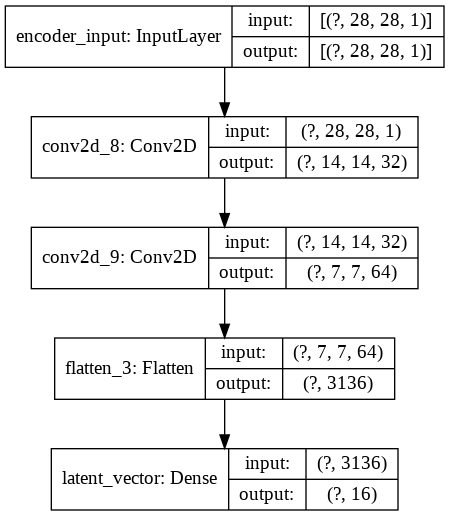}
		\caption{The encoder is comprised of two 2D convolution layers, one flatten layer, and one dense layer. This produces a lower dimension representation of the input data.}
		\label{ENC}
	\end{center}
\end{figure}

The CNN autoencoder is comprised of a convolutional encoder network, coupled with a decoding network. The encoder structure is shown in Figure.~\ref{ENC}.

\begin{figure}[h!]
	\begin{center}
		\includegraphics[width=.3\textwidth]{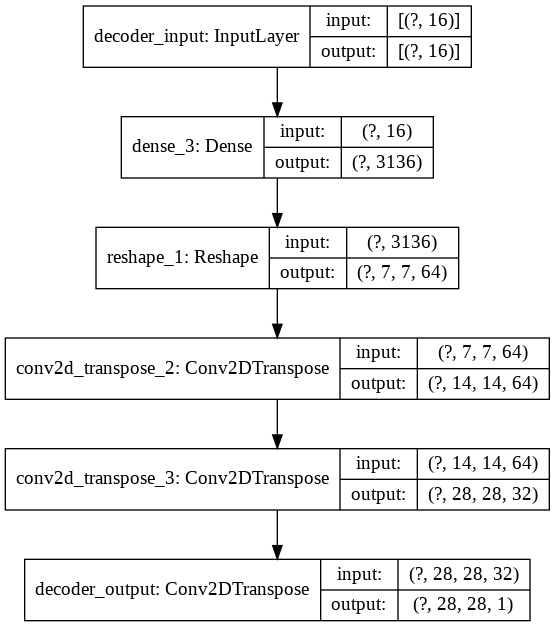}
		\caption{The decoder, which mirrors the encoder network. By matching the encoder's structure, the decoder can reproduce data structurally identical to the encoder input using the lower dimension representation created by the encoder. }
		\label{DEC}
	\end{center}
\end{figure}

\begin{figure}[h!]
	\begin{center}
		\includegraphics[width=.3\textwidth]{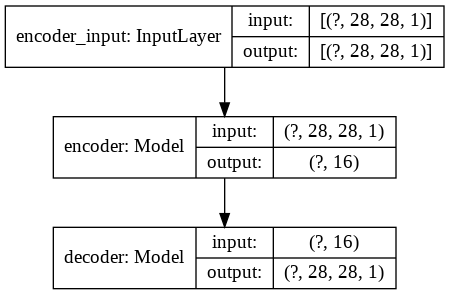} %
		\caption{The autoencoder, comprised of the encoder and decoder, allows the watchdog to generate input data based on the representations created at the waist layer.}
		\label{AES}
	\end{center}
\end{figure}

The decoding structure mirrors the encoder structure, as shown in Figure.~\ref{DEC}. The encoder and decoder structures are stacked to form the autoencoder. The resulting structure is shown in Figure.~\ref{AES}.

\subsubsection{Convolutional Neural Network Structure} 

\begin{figure}[h!]
	\begin{center}
		\includegraphics[width=.4\textwidth]{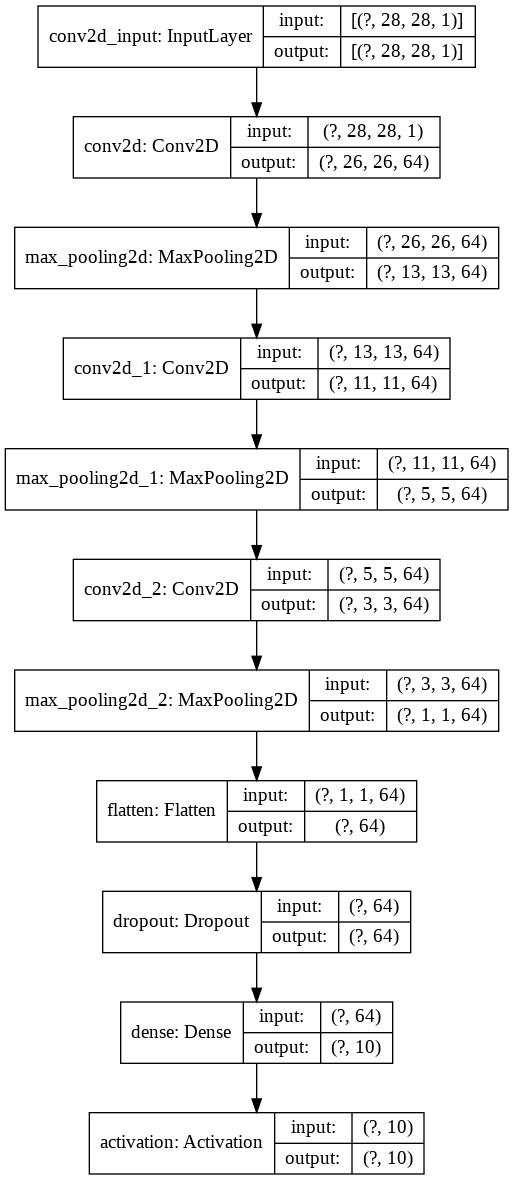} %
		\caption{Classification convolutional neural network structure, comprised of 3 Layers of 2D convolutions paired with 2D max pooling, one flatten layer, and one dropout layer, with a softmax activation layer.}
		\label{CNNS}
	\end{center}
\end{figure}

As demonstrated by Ciresan \cite{Ciresan}, Tabik \cite{Tabik}, and Bhatnagar \cite{Bhatnagar}, CNNs have shown impressive image classification capabilities. Our convolutional neural network classifier, described in Figure~\ref{CNNS}, is modeled after an example CNN provided by Geron in \cite{Geron}.

\subsection{Training the Networks}

\subsubsection{Training and evaluation datasets}
\begin{figure}[h]
	\centering
	\subfloat[An example of the MNIST digit 7]{
		\label{SDR}
		\includegraphics[width=.2\textwidth]{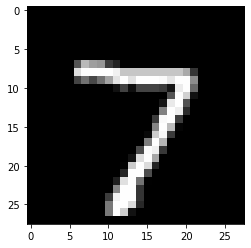}
	}
	\quad
	\subfloat[An example of the MNIST digit 4]{
		\label{FDR}
		\includegraphics[width=.2\textwidth]{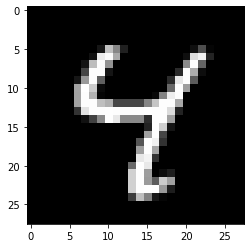}
	}
	\caption{MNIST digit image examples}
\end{figure}

With the structures of the networks established, we turn to identifying the training and evaluation datasets. The training data comes entirely from the MNIST handwritten digit dataset, and consists of 60,000 training and 10,000 test images evenly split across 10 classes of digits, 0-9. The evaluation dataset is augmented to include the fashion MNIST dataset test images. First introduced by Xiao in 2017 \cite{Xiao}, the fashion MNIST dataset is comprised of 70,000 total images, evenly distributed across 10 classes of different clothing types. These datasets were chosen due to their identical size, allowing for their easy use in the training and testing of both the autoencoder and classifier without modification. Both the autoencoder and the CNN were trained on  50,000 digit image dataset and validated on an additional 10,000 digit images from the training set. Examples of the training data are shown in Figures.~\ref{SDR} and \ref{FDR}. 

\begin{figure}[h]
	\centering
	\subfloat[An example of the fashion MNIST jacket class]{
		\label{JR}
		\includegraphics[width=.2\textwidth]{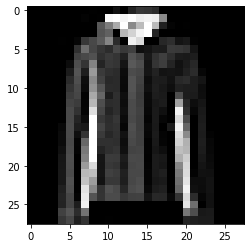}
	}
	\quad
	\subfloat[An example of the fashion MNIST pants class]{
		\label{PR}
		\includegraphics[width=.2\textwidth]{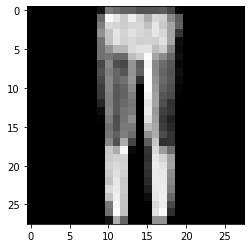}
	}
	\caption{Fashion MNIST image examples}
\end{figure}

In order to evaluate the effectiveness of the classifier and its watchdog, three evaluation datasets are used. The evaluation datasets are the combination of test images from the digit and fashion image sets. Additional examples can be seen in Figures.~\ref{JR} and \ref{PR}. Note that the evaluation images are separate from the training and validation datasets. The three datasets are as follows: in-distribution (digit images), out-of-distribution (fashion images), and mixed-distribution (both digit and fashion images).

\begin{enumerate}
	\item 10,000 test images from the MNIST digit dataset, in-distribution data
	\item 10,000 test images from the fashion MNIST dataset, out-of-distribution data
	\item 20,000 test images resulting from the combination of the MNIST digit and fashion MNIST test datasets, mixed-distribution data
\end{enumerate}


\section{Evaluating the Networks}

 \subsection{Evaluating the autoencoder}
\begin{figure}[h]
	\centering
	\subfloat[Watchdog regeneration of the in-distribution digit 7]{
		\label{SAE}
		\includegraphics[width=.2\textwidth]{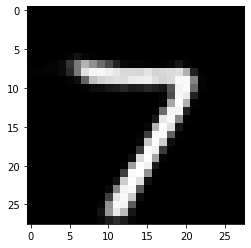}
	}
	\quad
	\subfloat[Watchdog regeneration of the in-distribution digit 4]{
		\label{FAE}
		\includegraphics[width=.2\textwidth]{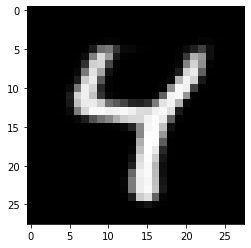}
	}
	\caption{Watchdog autoencoder regeneration of the in-distribution MNIST digit images}
\end{figure}

\begin{figure}[h]
	\centering
	\subfloat[Watchdog regeneration of the out-of-distribution jacket image.]{
		\label{JA}
		\includegraphics[width=.2\textwidth]{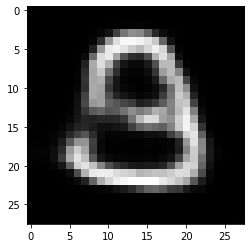} 
	}
	\quad
	\subfloat[Watchdog regeneration of the out-of-distribution pants image]{
		\label{PA}
		\includegraphics[width=.2\textwidth]{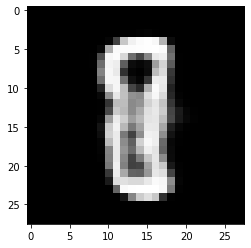}
	}
	\caption{Watchdog autoencoder regeneration of the out-of-distribution fashion MNIST images}
\end{figure}

 With the three evaluation datasets established, the performance of the autoencoder is examined. The MNIST digit and fashion MNIST datasets are passed through the autoencoder independently. The outputs of the autoencoder, examples of which can be seen in Figures.~\ref{SAE}, \ref{FAE}, \ref{JA}, and \ref{PA}, were then stored separately for additional analysis. The resulting generated images were then compared to their respective original images and the RMSE was calculated. In order to determine the range of values expected when calculating the RMSE, multiply the image size, 28x28x1 pixels, by the maximum pixel value, which has been normalized to values between 0 and 1. For this dataset, the range of RMSE values is between 0 and 28, with a RMSE of 0 representing a perfect match, and a RMSE of 28 representing a perfect mismatch.  Based on our experimentation, the average RMSE value calculated for the MNIST digit dataset was approximately 2.4, and the average RMSE value calculated for the fashion MNIST dataset was approximately 7.9. 
 
\subsection{Performance of the watchdog}

\subsubsection{ROC curves and classification errors}
In order to show the effectiveness of the watchdog, we produce receiver operator characteristic (ROC) curves. These curve show the tradeoff between the true positive vs. false positive rates. The rates are determined as: \[TPR = TP/(TP + FN)\] \[FPR = FP/(FP + TN)\], where: 

\begin{itemize}
	\item TP - True Positive = correct classification, the in-distribution inputs that are within the acceptance threshold
	\item FP - False Positive = incorrect classification, the out-of-distribution inputs are within the acceptance threshold
	\item FN - False Negative = incorrect classification, the out-of-distribution inputs are above the acceptance threshold
	\item TN - True Negative = correct classification, the in-distribution inputs are above the acceptance threshold
\end{itemize}

\begin{figure}[h]
	\centering
	\includegraphics[width=.45\textwidth]{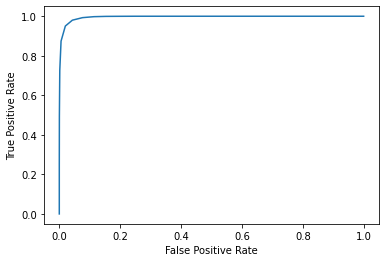}
	\caption{The ROC plot showing the performance of the watchdog using the mixed-distribution dataset. This curve has been produced based on the watchdog's ability to differentiate between in-distribution and out-of-distribution data as a function of RMSE threshold.}
	\label{AEROC}
\end{figure}

With the average RMSE values established, the next step is determining an appropriate threshold. Figure.~\ref{AEROC} shows the ROC curve for the watchdog autoencoder. This curve is the evaluation of the watchdog autoencoder based on its ability to separate the in-distribution digit images, or true positives, from out-of-distribution fashion images, or false positives.

\subsubsection{Monitoring the classifier with the watchdog autoencoder}
\begin{figure}
	\centering
	\includegraphics[width=.45\textwidth]{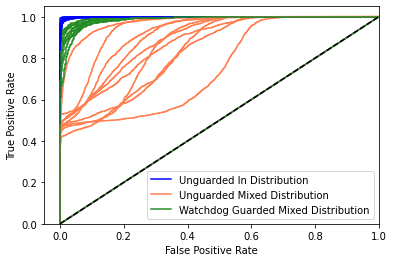}
	\caption{The ROC plots showing the watchdog performance on the three evaluation datasets. Blue indicates unguarded in-distribution performance, Orange indicates unguarded mixed-distribution performance, and Green indicates guarded mixed-distribution performance.}
	\label{WDG}
\end{figure}

The value of adding the autoencoder watchdog to the mixed-distribution dataset can be seen in Figure.~\ref{WDG} where the guarded mixed-distribution dataset has better performance than the unguarded mixed-distribution. As a point of reference, the ideal scenario, data contained only in the  in-distribution dataset, has been included in Figure.~\ref{WDG}. As we have shown, the results from the watchdog produce a more accurate true positive vs. false positive rate, and a more vertical ROC curve, when compared to the individual dataset performance.

\begin{figure}
	\centering
	\includegraphics[width=.45\textwidth]{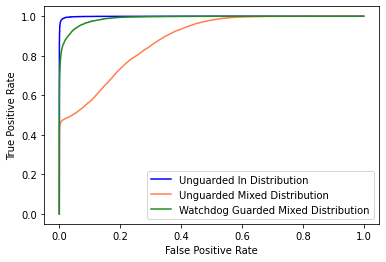}
	\caption{The averaged ROC plots of the unguarded in-distribution, unguarded mixed-distribution, and guarded mixed-distribution plots, as seen in Figure.~\ref{WDG} above.}
	\label{WDG_avg}
\end{figure}

\begin{figure}
	\centering
	\includegraphics[width=.45\textwidth]{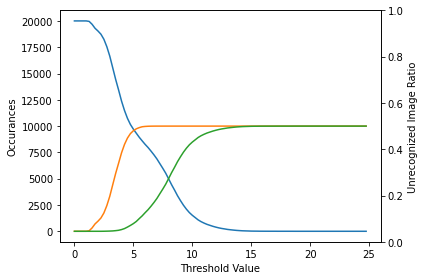}
	\caption{The distribution of images as a function of RMSE threshold. Blue represents unrecognized images (images that exceed RMSE threshold), orange represents in-distribution images (True Positives), and green represents out-of-distribution images (False Positives)}
	\label{GBJ}
\end{figure}

Along with the ROC curves, an interesting metric to note is the number of unrecognized images that have been detected in the dataset. The number of unrecognized images and the unrecognized image ratio, as seen in Figure.~\ref{GBJ}, can be used as tools to help determine a final threshold value when designing and developing watchdog guarded networks.

\section{Conclusion}

An initial proof of concept neural network watchdog is proposed to help improve the performance of classifiers on various datasets. The approach is also transparently applicable to regression neural networks. The choice of RMSE threshold is ultimately determined by the desired detection versus false alarm tradeoff. Alternately, the RMSE can be used to inform users of a measure of closeness of an input to the manifold of the watchdog autoencoder defined in-distribution manifold in feature space.


\begin{thebibliography}{00}

\bibitem{Abbasi} Abbasi, Mahdieh, et al. "Toward Metrics for Differentiating Out-of-Distribution Sets." arXiv preprint arXiv:1910.08650 (2019).

\bibitem{Alain} Alain, Guillaume, and Yoshua Bengio. "Understanding intermediate layers using linear classifier probes." arXiv preprint arXiv:1610.01644 (2016).

\bibitem{Alvernaz} Alvernaz, Samuel, and Julian Togelius. "Autoencoder-augmented neuroevolution for visual doom playing." 2017 IEEE Conference on Computational Intelligence and Games (CIG). IEEE, 2017.

\bibitem{Baur} Baur, Christoph, et al. "Deep autoencoding models for unsupervised anomaly segmentation in brain MR images." International MICCAI Brainlesion Workshop. Springer, Cham, 2018.

\bibitem {Bhatnagar} Bhatnagar, Shobhit, Deepanway Ghosal, and Maheshkumar H. Kolekar. "Classification of fashion article images using convolutional neural networks." 2017 Fourth International Conference on Image Information Processing (ICIIP). IEEE, 2017.

\bibitem{Ciresan} Ciresan, Dan Claudiu, et al. "Flexible, high performance convolutional neural networks for image classification." Twenty-Second International Joint Conference on Artificial Intelligence. 2011.

\bibitem{DeVries} DeVries, Terrance, and Graham W. Taylor. "Learning confidence for out-of-distribution detection in neural networks." arXiv preprint arXiv:1802.04865 (2018).

\bibitem{Geron} Géron, Aurélien. Hands-On Machine Learning with Scikit-Learn, Keras, and TensorFlow: Concepts, Tools, and Techniques to Build Intelligent Systems. O'Reilly Media, 2019.

\bibitem{Gondara} Gondara, Lovedeep. "Medical image denoising using convolutional denoising autoencoders." 2016 IEEE 16th International Conference on Data Mining Workshops (ICDMW). IEEE, 2016.

\bibitem{Guttormsson} Guttormsson, Sigurour E., R. J. Marks, M. A. El-Sharkawi, and I. Kerszenbaum. "Elliptical novelty grouping for on-line short-turn detection of excited running rotors." IEEE Transactions on Energy Conversion 14, no. 1 (1999): 16-22.

\bibitem{Huang} Huang, Huaibo, et al. "Introvae: Introspective variational autoencoders for photographic image synthesis." Advances in neural information processing systems. 2018.

\bibitem{Larsen} Larsen, Anders Boesen Lindbo, et al. "Autoencoding beyond pixels using a learned similarity metric." International conference on machine learning. 2016.

\bibitem{LeCun} LeCun, Yann, Yoshua Bengio, and Geoffrey Hinton. "Deep learning." nature 521.7553 (2015): 436-444.

\bibitem{Lore} Lore, Kin Gwn, Adedotun Akintayo, and Soumik Sarkar. "LLNet: A deep autoencoder approach to natural low-light image enhancement." Pattern Recognition 61 (2017): 650-662.

\bibitem{Lu} Lu, Xugang, et al. "Speech enhancement based on deep denoising autoencoder." Interspeech. 2013.

\bibitem{Luo} Luo, Junyu, et al. "Learning inverse mapping by autoencoder based generative adversarial nets." International Conference on Neural Information Processing. Springer, Cham, 2017.

\bibitem{Marchi} Marchi, Erik, et al. "A novel approach for automatic acoustic novelty detection using a denoising autoencoder with bidirectional LSTM neural networks." 2015 IEEE international conference on acoustics, speech and signal processing (ICASSP). IEEE, 2015.

\bibitem{Meng} Meng, Qinxue, et al. "Relational autoencoder for feature extraction." 2017 International Joint Conference on Neural Networks (IJCNN). IEEE, 2017.

\bibitem{Mescheder} Mescheder, Lars, Sebastian Nowozin, and Andreas Geiger. "Adversarial variational bayes: Unifying variational autoencoders and generative adversarial networks." Proceedings of the 34th International Conference on Machine Learning-Volume 70. JMLR. org, 2017.

\bibitem{Ng} Ng, Andrew. "Sparse autoencoder." CS294A Lecture notes 72.2011 (2011): 1-19.

\bibitem{Markopoulos} Markopoulos, Panos P., et al. "Efficient L1-norm principal-component analysis via bit flipping." IEEE Transactions on Signal Processing 65.16 (2017): 4252-4264.

\bibitem{Martin} Martin-Clemente, Ruben, and Vicente Zarzoso. "On the link between L1-PCA and ICA." IEEE transactions on pattern analysis and machine intelligence 39.3 (2016): 515-528.

\bibitem{Oja} Oja, Erkki. "Neural networks, principal components, and subspaces." International journal of neural systems 1, no. 01 (1989): 61-68.

\bibitem{Qi} Qi, Yumei, et al. "Stacked sparse autoencoder-based deep network for fault diagnosis of rotating machinery." Ieee Access 5 (2017): 15066-15079.

\bibitem{Radford} Radford, Alec, Luke Metz, and Soumith Chintala. "Unsupervised representation learning with deep convolutional generative adversarial networks." arXiv preprint arXiv:1511.06434 (2015).

\bibitem{Ranjan} Ranjan, Anurag, et al. "Generating 3D faces using convolutional mesh autoencoders." Proceedings of the European Conference on Computer Vision (ECCV). 2018.

\bibitem{Reed} Reed, Russell, and Robert J. Marks II. {\em Neural smithing: supervised learning in feedforward artificial neural networks}. MIT Press, 1999.

\bibitem{Rolfe} Rolfe, Jason Tyler. "Discrete variational autoencoders." arXiv preprint arXiv:1609.02200 (2016).

\bibitem{Shwartz} Shwartz-Ziv, Ravid, and Naftali Tishby. "Opening the black box of deep neural networks via information." arXiv preprint arXiv:1703.00810 (2017).

\bibitem{Streifel} Streifel, Robert J., R. J. Marks, M. A. El-Sharkawi, and I. Kerszenbaum. "Detection of shorted-turns in the field winding of turbine-generator rotors using novelty detectors-development and field test." IEEE Transactions on Energy Conversion 11, no. 2 (1996): 312-317.   

\bibitem{Tabik} Tabik, Siham, et al. "A snapshot of image pre-processing for convolutional neural networks: case study of MNIST." International Journal of Computational Intelligence Systems 10.1 (2017): 555-568.

\bibitem{Thompson} Thompson, Benjamin, Robert J. Marks, Jai J. Choi, Mohamed A. El-Sharkawi, Ming-Yuh Huang, and Carl Bunje. "Implicit learning in autoencoder novelty assessment." In Proceedings of the 2002 International Joint Conference on Neural Networks. IJCNN'02 (Cat. No. 02CH37290), vol. 3, pp. 2878-2883. IEEE, 2002.

\bibitem{Thompson2} Thompson, Benjamin B., R. J. Marks, and Mohamed A. El-Sharkawi. "On the contractive nature of autoencoders: Application to missing sensor restoration." In Proceedings of the International Joint Conference on Neural Networks, 2003., vol. 4, pp. 3011-3016. IEEE, 2003.

\bibitem{Tountas} Tountas, Konstantinos, Dimitris A. Pados, and Michael J. Medley. "Conformity evaluation and L1-norm principal-component analysis of tensor data." Big Data: Learning, Analytics, and Applications. Vol. 10989. International Society for Optics and Photonics, 2019.

\bibitem{Vincent} Vincent, Pascal, et al. "Extracting and composing robust features with denoising autoencoders." Proceedings of the 25th international conference on Machine learning. 2008.

\bibitem{Vu} Vu, Ha Son, et al. "Anomaly detection with adversarial dual autoencoders." arXiv preprint arXiv:1902.06924 (2019).

\bibitem{Wang} Wang, Yasi, Hongxun Yao, and Sicheng Zhao. "Auto-encoder based dimensionality reduction." Neurocomputing 184 (2016): 232-242.

\bibitem{Xia} Xia, Rui, et al. "Modeling gender information for emotion recognition using denoising autoencoder." 2014 IEEE International Conference on Acoustics, Speech and Signal Processing (ICASSP). IEEE, 2014.

\bibitem{Xiao} Xiao, Han, Kashif Rasul, and Roland Vollgraf. "Fashion-mnist: a novel image dataset for benchmarking machine learning algorithms." arXiv preprint arXiv:1708.07747 (2017).

\bibitem{Yadav} Yadav, Neha, Anupam Yadav, and Manoj Kumar. "History of neural networks." An Introduction to Neural Network Methods for Differential Equations. Springer, Dordrecht, 2015. 13-15.

\bibitem {Yu} Yu, Yang, Jun Long, and Zhiping Cai. "Network intrusion detection through stacking dilated convolutional autoencoders." Security and Communication Networks 2017 (2017).

\bibitem{Zeiler} Zeiler, Matthew D., and Rob Fergus. "Visualizing and understanding convolutional networks." European conference on computer vision. Springer, Cham, 2014.

\bibitem{Zhang} Zhang, Xiang, and Yann LeCun. "Text understanding from scratch." arXiv preprint arXiv:1502.01710 (2015).

\bibitem{Zhang2} Zhang, Zhifei, Yang Song, and Hairong Qi. "Gans powered by autoencoding a theoretic reasoning." ICML Workshop on Implicit Models. 2017.

\bibitem{Zhang3} Zhang, Zijun, et al. "Perceptual generative autoencoders." arXiv preprint arXiv:1906.10335 (2019).

\end{thebibliography}
\end{document}